\documentclass{article}
\usepackage{arxiv}
\usepackage[utf8]{inputenc} % allow utf-8 input
\usepackage[T1]{fontenc}    % use 8-bit T1 fonts
\usepackage{hyperref}       % hyperlinks
\usepackage{url}            % simple URL typesetting
\usepackage{booktabs}       % professional-quality tables
\usepackage{amsfonts}       % blackboard math symbols
\usepackage{graphicx}
\usepackage{natbib}
\usepackage{doi}
\usepackage{caption}
\usepackage{subcaption}

\title{"Hey..! This medicine made me sick": Sentiment Analysis of User-Generated Drug Reviews using Machine Learning Techniques}

\author{Abhiram B. Nair\\
	School of Digital Sciences\\
	Kerala University of Digital Sciences-\\Innovation and Technology\\
	Thiruvananthapuram, India \\
	\texttt{abhiram.ds22@duk.ac.in} \\
        \And
        Abhinand K.\\
	School of Digital Sciences\\
	Kerala University of Digital Sciences-\\Innovation and Technology\\
	Thiruvananthapuram, India \\
	\texttt{abhinand.ds22@duk.ac.in} \\
	\And
        Anamika U.\\
	School of Digital Sciences\\
	Kerala University of Digital Sciences-\\Innovation and Technology\\
	Thiruvananthapuram, India \\
	\texttt{anamika.ds22@duk.ac.in} \\
        \And
       Denil Tom Jaison\\
	School of Digital Sciences\\
	Kerala University of Digital Sciences-\\Innovation and Technology\\
	Thiruvananthapuram, India \\
	\texttt{deniljaison.ds22@duk.ac.in} \\
	\And
    Ajitha V.\\
	School of Digital Sciences\\
	Kerala University of Digital Sciences-\\Innovation and Technology\\
	Thiruvananthapuram, India \\
	\texttt{ajitha.ds22@duk.ac.in} \\
    \And
	V. S. Anoop \\
	School of Digital Sciences\\
	Kerala University of Digital Sciences-\\Innovation and Technology\\
	Thiruvananthapuram, India \\
	\texttt{anoop.vs@duk.ac.in} \\
}

\date{}

\renewcommand{\shorttitle}{Sentiment Analysis of User-Generated Drug Reviews using Machine Learning Techniques}

\hypersetup{
}

\begin{document}
\maketitle

\begin{abstract}
Sentiment analysis has become increasingly important in healthcare, especially in the biomedical and pharmaceutical fields. The data generated by the general public on the effectiveness, side effects, and adverse drug reactions are goldmines for different agencies and medicine producers to understand the concerns and reactions of people. Despite the challenge of obtaining datasets on drug-related problems, sentiment analysis on this topic would be a significant boon to the field. This project proposes a drug review classification system that classifies user reviews on a particular drug into different classes, such as positive, negative, and neutral. This approach uses a dataset that is collected from publicly available sources containing drug reviews, such as drugs.com. The collected data is manually labeled and verified manually to ensure that the labels are correct. Three pre-trained language models, such as BERT, SciBERT, and BioBERT, are used to obtain embeddings, which were later used as features to different machine learning classifiers such as decision trees, support vector machines, random forests, and also deep learning algorithms such as recurrent neural networks. The performance of these classifiers is quantified using precision, recall, and f1-score, and the results show that the proposed approaches are useful in analyzing the sentiments of people on different drugs.
\end{abstract}

% keywords can be removed

\section{Introduction}
In the vast expanse of healthcare, the experiences and opinions of individuals play a pivotal role in shaping the landscape of drug development, usage, and overall patient care\cite{hoffman2024artificial}. With the advent of user-generated content on online platforms, a wealth of valuable information is at our disposal. The proliferation of online forums, social media, and dedicated platforms for health-related discussions has given rise to an unprecedented volume of user-generated content, particularly in the form of drug reviews\cite{haque2023improving}\cite{basiri2020novel}. Leveraging the power of sentiment analysis and machine learning methodologies presents an opportunity to distill meaningful insights from this vast sea of unstructured data\cite{anoop2023sentiment}. Sentiment analysis, with its ability to provide insights acquired from people's experiences and emotions, has reshaped a number of enterprises\cite{jickson2023machine}\cite{john2023health}. Sentiment analysis helps identify sentiments expressed by individuals regarding the effectiveness, side effects, experiences, and overall satisfaction with specific drugs. Healthcare providers and pharmaceutical companies leverage sentiment analysis to gain insights into how patients perceive and experience certain medications. This aids in understanding real-world effectiveness, patient satisfaction, and potential areas for improvement. Understanding patient experiences is paramount in healthcare and pharmaceutical industries due to the direct impact on lives. Beyond mere feedback, these experiences inform patient-centered care, enabling tailored treatments and better outcomes. They uncover nuances in medications' efficacy, safety, and side effects, guiding quality improvement.

In recent years, exponential growth has been witnessed in the amount of user-generated data added to social media platforms such as Facebook and discussion forums such as drugs.com and Reddit\cite{anoop2019aspect}\cite{varghese2022deep}\cite{lekshmi2022sentiment}. People share their opinions, feedback, and any concerns on these platforms enabling several discussions and deliberations\cite{anoop2023we}\cite{anoop2023public}. This is also applicable in the context of drugs and medicines, where users may share their experiences using the medicine, critical reviews, pharmacy experiences, and adverse drug reactions. This information will be highly crucial and may be a goldmine for companies and other stakeholders to understand user perceptions, discourse, and sentiments of a particular drug in the market. This is a well-studied problem, and many approaches have been reported in the literature in the recent past, with varying degrees of success. With the advancements in natural language processing and machine learning techniques, pre-trained language models become popular and show state-of-the-art performances compared with older approaches. In this connection, this paper proposes an approach for the sentiment classification of drug reviews generated by the users and publicly available. We use advanced pre-trained models such as BERT, BioBERT, and SciBERT for better feature encoding that may improve the sentiment classification task. The major contributions of this paper are as follows:
\begin{itemize}
    \item Discusses the relevance and state-of-the-art approaches in drug review classification using natural language processing and machine learning techniques
    \item Proposed an approach for sentiment classification of publicly available drug reviews using machine learning approaches with different pretrained models such as BERT, BioBERT, and SciBERT
    \item Compare the performance of different machine learning algorithms for dug review sentiment classification in terms of precision, recall, and f-measure
\end{itemize}

\section{Related Studies}
This section discusses some of the approaches already reported in the sentiment classification literature, specifically on drug and patient reviews that are generated by the public. Satvik Garg et al. suggest a creative solution to the medical resource crisis by developing a pharmaceutical recommendation system\cite{garg2021drug}. The system uses patient reviews to anticipate attitudes and prescribe the best treatment for a given ailment, leveraging machine learning algorithms, sentiment analysis, and feature engineering. The findings demonstrate the usefulness of the LinearSVC classifier with TF-IDF vectorization, which achieves a remarkable 93\% accuracy while beating other models. The study digs into the complexities of the medicine recommendation process, examining several vectorization algorithms used for sentiment analysis as well as the evaluation metrics used to assess classifier effectiveness. While giving promising results, the report openly admits the framework's limits and recommends ways to enhance it in the future, contributing to the ongoing discussion about optimizing medical resource allocation. The paper by B. Lokeswara Nayak et al. introduces a drug recommendation system based on sentiment analysis of patient reviews. The system utilizes Bow, TF-IDF, Word2Vec, and Manual Feature Analysis, with the Linear SVC classifier using TF-IDF achieving the highest accuracy of 93\%. Motivated by doctors' challenges in treatment decisions, the system aims to reduce medication errors and enhance patient well-being by providing accurate information on drug effectiveness. The study highlights the system's potential to empower users in making informed health-related decisions, improve healthcare provider practices, and contribute to overall patient care.

In the study authored by GV Lavanya et al. \cite{lavanyadrug}, the drug recommender system is meticulously designed to employ machine learning for sentiment analysis, creating a precise medicine recommendation system. Utilizing a diverse dataset sourced from reputable outlets, the system integrates sentiment analysis with drug recommendation algorithms. The approach encompasses collaborative filtering, content-based filtering, and hybrid filtering methods to enhance the accuracy and reliability of drug recommendations. The LinearSVC classifier, employing TF-IDF vectorization, emerges as the top-performing model with an impressive accuracy rate of 93\%. The system rates drugs and gauges public opinion, optimizing medicine selection to improve patient outcomes and reduce adverse reactions. Furthermore, the study expands its focus to recommend diabetes drugs through case-based reasoning (CBR) and the Nearest Neighbors approach. By comparing current issues with past cases and assessing similarities, the system suggests tailored treatment plans for individuals based on their specific diabetes subtype. This comprehensive approach holds significant promise in revolutionizing personalized medicine recommendations. In the paper by Mohapatra et al. \cite{mohapatra2022machine} introduces a machine learning-based drug recommendation system for healthcare. It employs content-based filtering and collaborative filtering methodologies to suggest drugs based on input conditions. Content-based filtering relies on the similarity of input conditions to others in the dataset, while collaborative filtering considers correlations with similar conditions. Leveraging machine learning and data mining, the system analyzes large datasets for accurate drug recommendations. The author emphasizes the importance of data analysis and visualization for better comprehension of drug review datasets. The system aims to assist healthcare professionals in making informed prescription decisions by considering conditions, ratings, and reviews. The user-item matrix forms the basis for collaborative filtering, with the system sorting values by rating to recommend suitable medicines based on correlation with similar conditions.

Another approach was reported to assess drug effectiveness using five machine-learning algorithms applied to drug reviews. They employ a classification approach, categorizing drugs into five classes based on effectiveness.\cite{uddin2022drug} Precision, recall, and f1-score metrics are utilized to evaluate algorithm performance. Notably, the Random Forest algorithm outperforms the other four, demonstrating superior accuracy, precision, recall, and f1-score for effective drugs compared to ineffective ones. The f1-score, chosen for its balance between precision and recall, offers a nuanced accuracy perspective. This innovative research introduces a novel aspect by categorizing drugs into five effectiveness classes, providing valuable insights for consumers and manufacturers regarding a drug's efficacy and potential side effects. The paper introduces a novel methodology for sentiment analysis of drug reviews, leveraging SAS Enterprise Miner and WIT.AI. The approach employs text analytics, predictive models, and natural language processing to classify side effects and drug effectiveness.\cite{chauhansentiment} The use of sentiment lexicons aids in identifying opinion terms and trends in medical texts. The authors address the challenge of unstructured drug reviews and employ WIT.AI to determine average sentiment and extract key positive and negative reviews. The methodology has the potential to transform healthcare decision-making for customers, offering actionable insights for healthcare professionals. Noteworthy for its integration of diverse techniques, the paper stands out in the sentiment analysis of drug reviews.

The paper introduces a novel model leveraging natural language processing (NLP) and machine learning algorithms to predict drug ratings and identify drugs causing adverse reactions through review analysis.\cite{gawich2022proposed} The model was applied to the Drugs.com dataset and evaluated on the independent drugs.lib dataset demonstrates high accuracy. Key to its success is a hybrid approach employing the Weakly Supervised Mechanism (WSM) and a Convolutional Neural Network (CNN) with bi-directional Long-Short Term Memory (LSTM) for classifying reviews into positive or negative ratings. Sentiment extraction, facilitated by the syuzhet package in the R-program, aids in discovering drugs provoking adverse reactions by analyzing emotional nuances in reviews. The model's innovation lies in its comprehensive analysis of patient experiences, addressing challenges like slang, sentiment expressions, and poor grammar in social media reviews. The proposed model emerges as a valuable tool for the medical and pharmaceutical sectors, shedding light on patients' perspectives and guiding informed decision-making. The paper introduces a drug recommendation system employing sentiment analysis in drug reviews through machine learning. \cite{hossain2020drugs}Its goal is to provide valuable public opinion summaries for patients, pharmacists, and clinicians. The study evaluates Decision Tree, K-Nearest Neighbors, and Linear Support Vector Classifier algorithms for rating generation, ultimately selecting the latter for a balanced trade-off between accuracy, efficiency, and scalability. Stressing the importance of sentiment analysis in healthcare decision-making, the proposed system framework consists of five modules: data preprocessing, rating generation, model evaluation, sentiment analysis, and recommendation modeling. Key features encompass a sentimental measurement approach, consideration of user usefulness and patient conditions, and a recommendation model. By factoring in drug reviews' utility, patient conditions, and sentiment polarity, the system facilitates improved drug selection, offers valuable public opinion summaries, enhances understanding of patient needs, supports decision-making, and contributes to overall public health improvement.

These studies underline that there are a lot of avenues where natural language processing and machine learning techniques can be utilized to their full potential for classifying the sentiments of drug reviews generated by the users to find interesting patterns and discourse. This may be immensely useful for different stakeholders in analyzing the user sentiments and trends that may aid in improving their products and services. The details of the proposed approach are discussed in detail in the subsequent sections.

\section{Materials and Methods}
This section highlights the materials and methods used in the proposed approach for implementing machine learning and natural language processing techniques. Different libraries used and also the different pre-trained models are discussed in this section.
\subsection{Beautful Soup}
Beautiful Soup is a Python library designed for web scraping, a process of extracting data from HTML and XML files on websites. Its primary purpose is to simplify the parsing and navigation of web page structures, creating a parse tree of Python objects that represents the document's hierarchy. Beautiful Soup provides convenient methods for searching and manipulating this tree, making it easier to locate specific tags, attributes, or text content within a web page.
\subsection{Bidirectional Encoder Representations from Transformers}
In 2018, Google introduced Bidirectional Encoder Representations from Transformers (BERT), which completely changed the field of natural language processing (NLP). As a cutting-edge model, it uses a bidirectional technique to examine words' left and right contexts in a sentence, improving its ability to comprehend contextual linkages. BERT uses self-attention mechanisms within the Transformer architecture to determine the relative relevance of words based on contextual relationships. The model can learn complex language patterns by performing tasks like predicting masked words and sentence associations because it has been pre-trained on large corpora that include a variety of text sources. BERT's remarkable performance is made possible by its contextual embeddings, which capture fine-grained syntactic and semantic structures. Interestingly, the open-source model has been widely adopted and fine-tuned for use in particular NLP tasks. The training corpora, which comprise a diverse range of textual material, enhance the adaptability of the model. Moreover, BERT uses a dynamic vocabulary built using WordPiece tokenization, which enables it to handle a variety of linguistic variances with flexibility. This versatility is crucial to BERT's continuous influence on the advancement of transformer-based models.
\subsection{SciBERT}
A specialized version of the BERT (Bidirectional Encoder Representations from Transformers) model that has been carefully designed for scientific text comprehension is called SciBERT. Designed to enhance efficiency on various natural language processing (NLP) tasks within the scientific domain, SciBERT was specifically trained on an extensive corpus sourced from academic papers on Semantic Scholar. SciBERT stands apart from BERT even though it retains the same basic design because it trains just on scientific texts. This focused approach guarantees that the model can accurately represent the complex linguistic structures found in scientific writing. Of note, SciBERT uses its unique "scivocab" wordpiece vocabulary, built on the scientific corpus and made possible by the SentencePiece library. By adding to the model's domain-specific adaptability, this specialized vocabulary helps it better express and understand scientific language and settings. In simple terms, SciBERT is an advanced natural language processing (NLP) model specifically designed for scientific communication. It provides enhanced language representation and contextual understanding in the scientific domain.
\subsection{BioBERT}
BioBERT is a modified version of the BERT (Bidirectional Encoder Representations from Transformers) that is developed for biomedical text processing. This version takes into account the special linguistic peculiarities and terminology used in the biomedical field. Although the model's basic design is similar to that of BERT, it has been pre-trained on an extensive corpus of biological literature, which includes clinical notes and PubMed articles. From a scientific perspective, BioBERT improves contextual embeddings by identifying the complex word associations seen in biomedical literature. BERT's bidirectional approach makes it possible to fully comprehend contextual dependencies. The model uses the Transformer architecture's self-attention mechanisms to dynamically determine a word's importance based on its contextual relationships. In particular, BioBERT leverages a specialized vocabulary called "biocides," which is built with WordPiece tokenization. This lexicon is customized for use in the biomedical field and includes a wide range of terminology unique to medical literature. This modification guarantees that the model is capable of managing the domain-specific vocabulary and terminology found in writings related to healthcare. Predicting masked words and sentence associations in the biomedical corpus is one of the pre-training tasks of BioBERT, which helps individuals acquire domain-specific language patterns. The model is further refined for specialized biomedical applications, like biomedical named entity recognition, relation extraction, and other biomedical information extraction tasks, by fine-tuning downstream tasks.
\subsection{SBERT}
A version of the BERT (Bidirectional Encoder Representations from Transformers) model designed for sentence-level tasks is called Sentence-BERT (S-BERT). S-BERT's primary breakthrough is in its ability to produce fixed-size sentence embeddings that accurately represent the semantic meaning of complete phrases. S-BERT is specially taught to grasp and represent whole sentences, whereas BERT is primarily trained for sentence-level tasks and masked word prediction. By implementing a Siamese network design during training, S-BERT accomplishes this. Given a pair of sentences as input, the Siamese network learns to produce dissimilar embeddings for sentences with diverse meanings and similar embeddings for semantically similar sentences. As a result, S-BERT can be applied to a number of sentence-level tasks, including clustering, sentence similarity, and paraphrase detection. Sentences in pairs are fed into the model during training, and the embeddings are optimized to bring sentences with similar meanings closer together in the embedding space. In this sense, S-BERT can be used for jobs where it is important to comprehend the overall semantic similarity or dissimilarity of phrases.
\subsection{Decision Tree} 
A decision tree is a machine learning predictive model that divides a dataset into a tree-like, hierarchical structure. The first step in the procedure is to choose the root node, which stands for the feature that best separates the data according to a given standard. In the case of classification, this is typically the Gini impurity, and in regression, the mean squared error. Each branch in the tree represents a feature test result, and internal nodes represent subsequent features and related conditions for data division. Recursive splitting keeps going until a stopping condition is satisfied, like a certain tree depth or a minimum number of samples per leaf. The final, predicted results are contained in the leaf nodes of the resulting tree, which offers a clear path for decision-making. Interpretability is a key component of decision trees, and different algorithms (e.g., ID3, CART) use statistical measures to identify the most informative features for splitting nodes. To reduce overfitting and improve the model's ability to generalise to new data, optional pruning strategies can be used after the model is constructed. In order to use the decision tree's predictive power, a new instance's feature values are used to guide a traversal of the tree from the root to a leaf node, which eventually yields a predicted class or value associated with that leaf node.
\subsection{Support Vector Classification}
Support Vector Classification (SVC) is a machine learning technique intended to categorize data points by finding the best hyperplane in a high-dimensional space. The hyperplane is positioned strategically to maximize the margin, which is the area between it and the nearest data points from different classes. This ideal separation is mostly determined by the support vectors or the data points that are closest to the hyperplane. By using kernel functions to translate input features into higher-dimensional spaces, SVC is skilled at managing non-linear interactions. The SVC decision function makes it easier to classify fresh data points by using feature weights and a bias term. A regularisation term provides a compromise between attaining a larger margin and permitting a certain amount of misclassification tolerance. The algorithm is trained by minimizing a cost function that penalizes misclassifications. The idea of a soft margin makes SVC an even more flexible and effective tool for classification tasks by increasing its capacity to adapt to datasets with intrinsic noise or incomplete separability.
\subsection{Random Forest}
Random Forest is an ensemble learning algorithm that aggregates the results of several decision trees to improve prediction accuracy. Using a method known as bootstrap aggregating (bagging), each tree in the ensemble is trained using a random subset of the training data. Notably, only a random subset of characteristics is taken into account at each node when building separate trees due to the application of feature randomization. A majority vote decides the final prediction for classification tasks, while an average is used for regression tasks. The decorrelation of trees—achieved by data and feature randomness—is the source of the algorithm's robustness and generalization abilities. Random Forest is a computationally efficient method since it allows for the parallel creation of individual trees. It does not require a separate validation set, but its capacity to estimate out-of-bag error offers a dependable indicator of generalization performance. All things considered, Random Forest is a strong and empirically supported machine learning method that excels at managing complicated datasets and increasing prediction accuracy.
\subsection{Recurrent Neural Network}
With the addition of recurrent connections, a Recurrent Neural Network (RNN) is a family of artificial neural networks created especially to process sequential input. The network can model temporal relationships in the input sequence owing to these connections, which enable the network to keep data from previous time steps in a concealed state. which enable the network to keep data from previous time steps in a concealed state. The hidden state acts as a kind of memory, influencing the network's predictions at each time step. RNNs are trained by the use of Backpropagation Through Time (BPTT), in which the network is unfolded over the temporal dimension and gradients are calculated at each time step. RNNs' inability to accurately capture long-range dependencies in sequential data is caused by their susceptibility to the vanishing gradient problem. Despite this disadvantage, RNNs are utilized in a wide range of applications, including speech recognition, natural language processing, and time series analysis, where the sequential character of the input is essential.
\section{Proposed Approach}
This section presents the proposed approach for sentiment classification of drug reviews using machine learning and natural language processing techniques. The overall workflow of the proposed approach is shown in Fig. 1.
\begin{figure}
	\centering
	\includegraphics[width=0.6\textwidth]{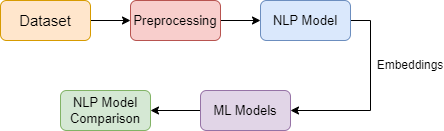}
	\caption{Our proposed model for sentiment analysis}
	\label{fig:fig4}
    \end{figure}

\begin{itemize}
    \item \textbf{Dataset}: The dataset used was built using 5,170 drug reviews taken from the publicly accessible website WebMD with the assistance of the Python library BeautifulSoup ,which was then manually categorized into three distinct groups based on their sentiment.The groups were namely  "Positive", "Neutral" and "Negative" thus resulting in a  dataset comprising two columns: "Reviews" and "Classification."
    \item \textbf{Preprocessing}: Data preprocessing is the crucial first step in data mining and machine learning, where raw and often messy data is transformed into a clean and organized format suitable for analysis and modeling. It involves a series of techniques to address inconsistencies, missing values, outliers, and errors, ensuring the data is accurate, consistent, and complete. Data preprocessing also includes data transformation, feature engineering, and data reduction to enhance its quality and usability for various algorithms.
    \item \textbf{Embeddings Generation}: Now we have the preprocessed data with reviews and labels, we have to transform these into numerical values. Hence, labels can be encoded using LabelEncoder, and embeddings of the reviews were created. Embeddings are low-dimensional representations of high-dimensional data, such as words, images, or audio. They capture the meaning and relationships between the data points, allowing machines to understand and process complex information. Embeddings are learned from large amounts of data and used in a wide range of applications, including natural language processing, recommender systems, image recognition, and anomaly detection. The reviews are inputted into various NLP models, including BERT, SciBERT, BioBERT, and S-BERT. The embeddings thus obtained from these respective models represent the reviews as vectors, capturing the meaning of the text by considering the context in which the words or sentences are in the review.
    \item \textbf{Machine Learning Models}: Upon generating the embeddings and identifying their corresponding labels, the classification process commences. This involves feeding the embeddings and their associated labels into various machine learning models. Commonly employed machine learning algorithms for classification tasks include decision tree classifiers, logistic regression, support vector classifiers (SVCs), and random forest classifiers. Each of these algorithms possesses unique strengths and limitations, and the choice of algorithm hinges on the specific characteristics of the data and the desired outcome.
    \item \textbf{Model Comparison}: The performance of each model is evaluated and compared.From the evaluation metrics, RNN is chosen for the model creation, which yields higher accuracy.
\end{itemize}

\section{Experimental Setup}
This project was done in a system with an Intel Core i5 processor having Windows 11. We have executed the code in Visual Studio code with Python environment 3.9.11. As collecting the data is the primary task, web scrapping should be done, followed by numerous other tasks. The Synthetic Minority Oversampling Technique (SMOTE) is implemented to avoid the problem of imbalance in the dataset. Now, there is a dataset with reviews and labels. The labels were encoded using LabelEncoder. Generated embeddings of the reviews using different pre-trained models like Bert, SciBERT, BioBERT, and SBERT using the Transformers library in Python. Later, the dataset is splitted for training and testing. Proper data splitting helps assess a model’s generalization performance and ability to make predictions on new and unseen data. Independant features were stored in one variable, and target values were saved in another. Several classification models, including Random Forest, Decision Tree, Support Vector Machine, and Logistic Regression, were applied to the dataset for analysis. From this, we should decide which one would be selected for our model. Random Forest have the best accuracy among the classifiers. A random forest classifier is an ensemble learning method that combines multiple decision trees to make predictions, reducing variance and improving accuracy.

\section{Results and Discussion}
This section presents the experiment's results with the proposed approach for sentiment analysis on user-generated drug reviews. The comparative results are presented in this section, followed by a detailed discussion of the results obtained. The precision, recall, and f1-score obtained for the different classifiers used in this work with BERT, SBERT, SciBERT, and BioBERT are given in Table 1, Table 2, Table 3, and Table 4, respectively. The training and testing accuracy for BERT and SBERT models for the decision tree, SVC, random forest, and logistic regression classifiers is shown in Fig. 2. The training and testing accuracy for BioBERT and SciBERT models for the decision tree, SVC, random forest, and logistic regression classifiers are shown in Fig. 3. The accuracy and loss for the BERT model for the decision tree, SVC, random forest, and logistic regression classifiers are shown in Fig. 4, the accuracy and loss for the SBERT model for decision tree, SVC, random forest, and logistic regression classifiers are shown in Fig. 5, and the accuracy and loss for the SciBERT model for decision tree, SVC, random forest, and logistic regression classifiers are shown in Fig. 6.
The weighted average for the precision, recall, and f1-score values for BERT with recurrent neural networks are 52\%, 53\%, and 50\%, respectively.
\begin{table}[htb]
\centering
\caption{The precision, recall, f1-score for different classifiers for BERT with Recurrent Neural Network}
\begin{tabular}{|r|r|r|r|r|}
\hline
                      & \textbf{Precision} & \textbf{Recall} & \textbf{F1-Score} & \textbf{Support} \\ \hline
\textbf{Negative}            & 0.61               & 0.68            & 0.64              & 506              \\ \hline
\textbf{Neutral}           & 0.38               & 0.8             & 0.13              & 201              \\ \hline
\textbf{Positive}            & 0.45               & 0.58            & 0.51              & 323              \\ \hline
\textbf{Accuracy}     &                    &                 & 0.53              & 1030             \\ \hline
\textbf{Macro Avg}    & 0.48               & 0.45            & 0.43              & 1030             \\ \hline
\textbf{Weighted Avg} & 0.52               & 0.53            & 0.50              & 1030             \\ \hline
\end{tabular}
\end{table}

\begin{table}[htb]
\centering
\caption{The precision, recall, f1-score for different classifiers for SBERT with Recurrent Neural Network}
\begin{tabular}{|r|r|r|r|r|}
\hline
                      & \textbf{Precision} & \textbf{Recall} & \textbf{F1-Score} & \textbf{Support} \\ \hline
\textbf{Negative}            & 0.61               & 0.72            & 0.66              & 506              \\ \hline
\textbf{Neutral}           & 0.26               & 0.20            & 0.23              & 201              \\ \hline
\textbf{Positive}            & 0.50               & 0.44            & 0.46              & 323              \\ \hline
\textbf{Accuracy}     &                    &                 & 0.53              & 1030             \\ \hline
\textbf{Macro Avg}    & 0.46               & 0.45            & 0.45              & 1030             \\ \hline
\textbf{Weighted Avg} & 0.51               & 0.53            & 0.51              & 1030             \\ \hline
\end{tabular}
\end{table}

\begin{table}[htb]
\centering
\caption{The precision, recall, f1-score for different classifiers for SciBERT with Recurrent Neural Network}
\begin{tabular}{|r|r|r|r|r|}
\hline
                      & \textbf{Precision} & \textbf{Recall} & \textbf{F1-Score} & \textbf{Support} \\ \hline
\textbf{Negative}            & 0.56               & 0.75            & 0.64              & 506              \\ \hline
\textbf{Neutral}           & 0.33               & 0.07            & 0.12              & 201              \\ \hline
\textbf{Positive}            & 0.47               & 0.46            & 0.46              & 323              \\ \hline
\textbf{Accuracy}     &                    &                 & 0.53              & 1030             \\ \hline
\textbf{Macro Avg}    & 0.46               & 0.43            & 0.41              & 1030             \\ \hline
\textbf{Weighted Avg} & 0.49               & 0.53            & 0.48              & 1030             \\ \hline
\end{tabular}
\end{table}

\begin{table}[htb]
\centering
\caption{The precision, recall, f1-score for different classifiers for BioBERT with Recurrent Neural Network}
\begin{tabular}{|r|r|r|r|r|}
\hline
                      & \textbf{Precision} & \textbf{Recall} & \textbf{F1-Score} & \textbf{Support} \\ \hline
\textbf{Negative}            & 0.60               & 0.71            & 0.65              & 506              \\ \hline
\textbf{Neutral}           & 0.30               & 0.10            & 0.16              & 201              \\ \hline
\textbf{Postive}            & 0.48               & 0.54            & 0.50              & 323              \\ \hline
\textbf{Accuracy}     &                    &                 & 0.54              & 1030             \\ \hline
\textbf{Macro Avg}    & 0.46               & 0.45            & 0.44              & 1030             \\ \hline
\textbf{Weighted Avg} & 0.50               & 0.54            & 0.51              & 1030             \\ \hline
\end{tabular}
\end{table}

\begin{figure}[ht!]
     \centering
     \begin{subfigure}[b]{0.48\textwidth}
         \centering
         \includegraphics[width=\textwidth]{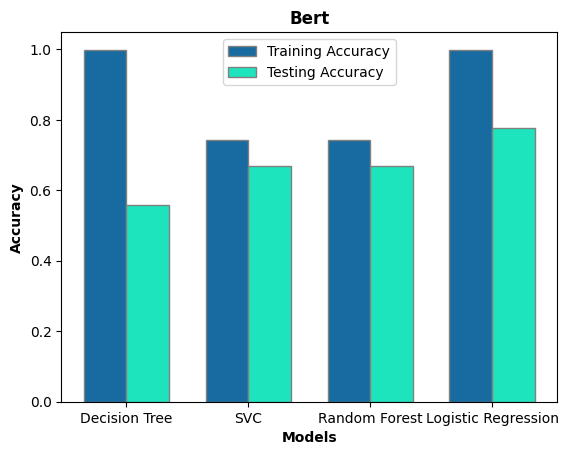}
         \caption{Training and testing accuracy for BERT model}
         \label{fig:y equals x}
     \end{subfigure}
     \hfill
     \begin{subfigure}[b]{0.48\textwidth}
         \centering
         \includegraphics[width=\textwidth]{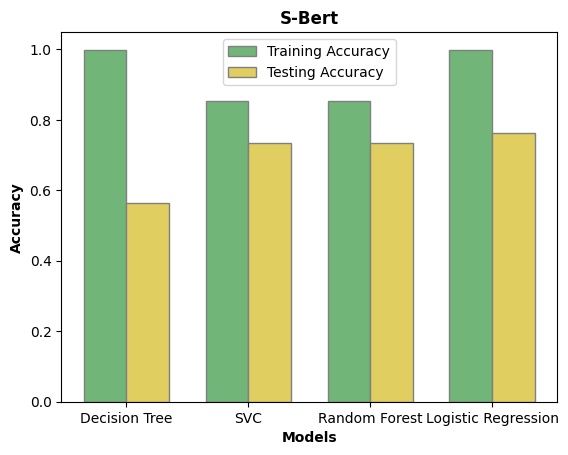}
         \caption{Training and testing accuracy for SBERT model}
         \label{fig:three sin x}
     \end{subfigure}
        \caption{Training and testing accuracy for BERT and SBERT models for decision tree, SVC, random forest, and logistic regression classifiers}
        \label{fig:three graphs}
\end{figure}

\begin{figure}[ht!]
     \centering
     \begin{subfigure}[b]{0.48\textwidth}
         \centering
         \includegraphics[width=\textwidth]{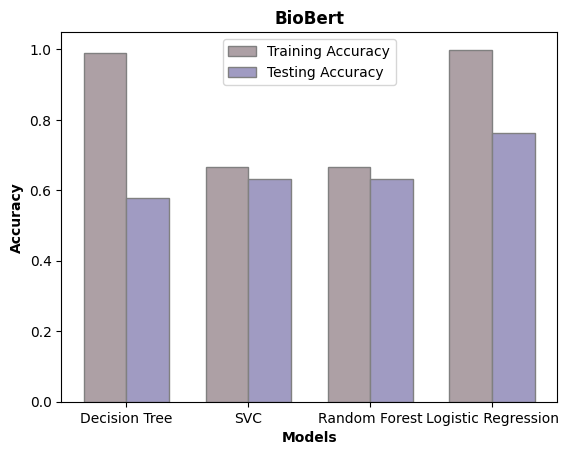}
         \caption{Training and testing accuracy for BioBERT model}
         \label{fig:y equals x}
     \end{subfigure}
     \hfill
     \begin{subfigure}[b]{0.48\textwidth}
         \centering
         \includegraphics[width=\textwidth]{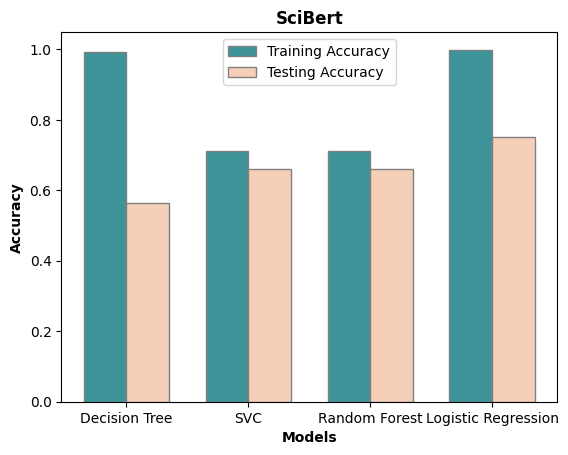}
         \caption{Training and testing accuracy for SciBERT model}
         \label{fig:three sin x}
     \end{subfigure}
        \caption{Training and testing accuracy for BioBERT and SciBERT models for decision tree, SVC, random forest, and logistic regression classifiers}
        \label{fig:three graphs}
\end{figure}

\begin{figure}[ht!]
     \centering
     \begin{subfigure}[b]{0.48\textwidth}
         \centering
         \includegraphics[width=\textwidth]{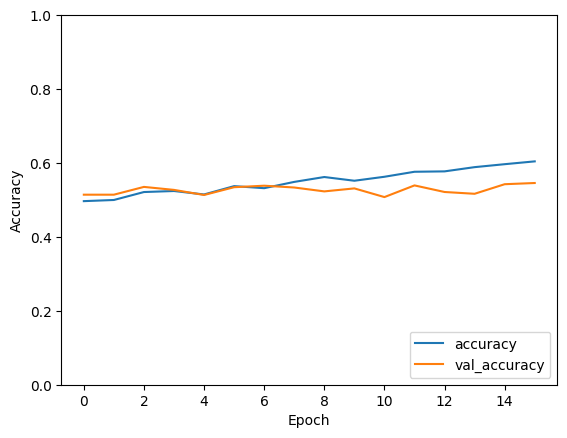}
         \caption{Accuracy for BERT model}
         \label{fig:y equals x}
     \end{subfigure}
     \hfill
     \begin{subfigure}[b]{0.48\textwidth}
         \centering
         \includegraphics[width=\textwidth]{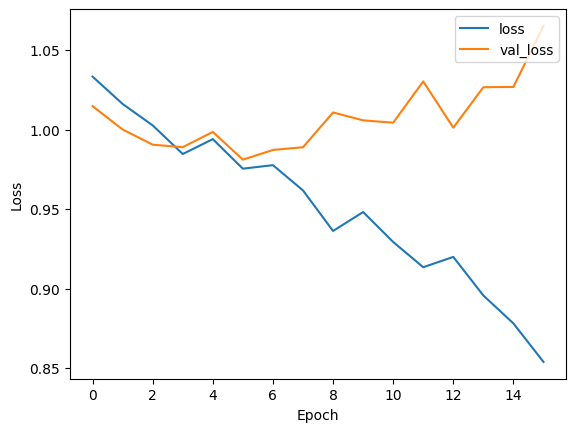}
         \caption{Loss for BERT model}
         \label{fig:three sin x}
     \end{subfigure}
        \caption{Accuracy and loss for the BERT model for decision tree, SVC, random forest, and logistic regression classifiers}
        \label{fig:three graphs}
\end{figure}

\begin{figure}[ht!]
     \centering
     \begin{subfigure}[b]{0.40\textwidth}
         \centering
         \includegraphics[width=\textwidth]{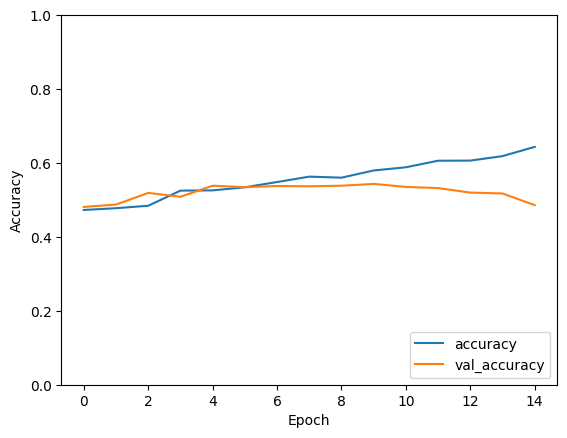}
         \caption{Accuracy for SBERT model}
         \label{fig:y equals x}
     \end{subfigure}
     \hfill
     \begin{subfigure}[b]{0.40\textwidth}
         \centering
         \includegraphics[width=\textwidth]{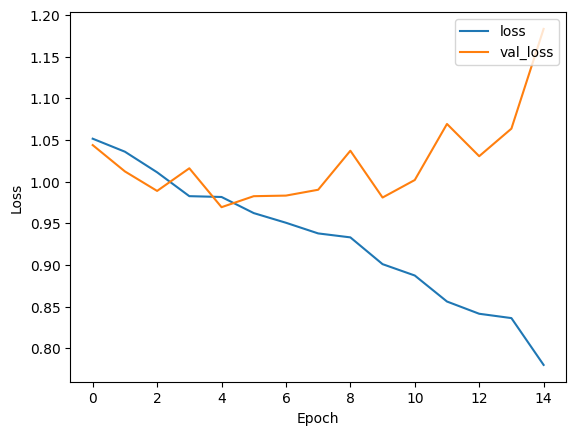}
         \caption{Loss for SBERT model}
         \label{fig:three sin x}
     \end{subfigure}
        \caption{Accuracy and loss for the SBERT model for decision tree, SVC, random forest, and logistic regression classifiers}
        \label{fig:three graphs}
\end{figure}
	
\begin{figure}[ht!]
     \centering
     \begin{subfigure}[b]{0.40\textwidth}
         \centering
         \includegraphics[width=\textwidth]{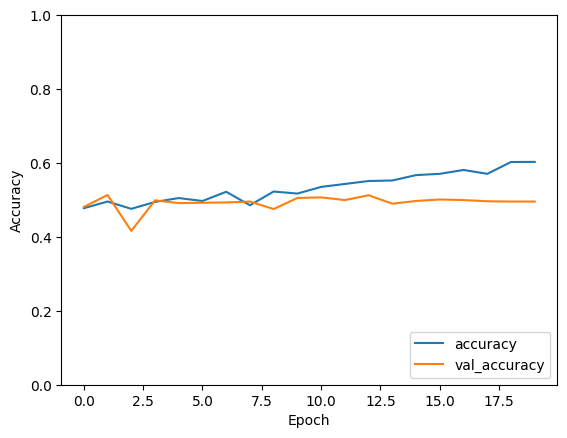}
         \caption{Accuracy for SciBERT model}
         \label{fig:y equals x}
     \end{subfigure}
     \hfill
     \begin{subfigure}[b]{0.40\textwidth}
         \centering
         \includegraphics[width=\textwidth]{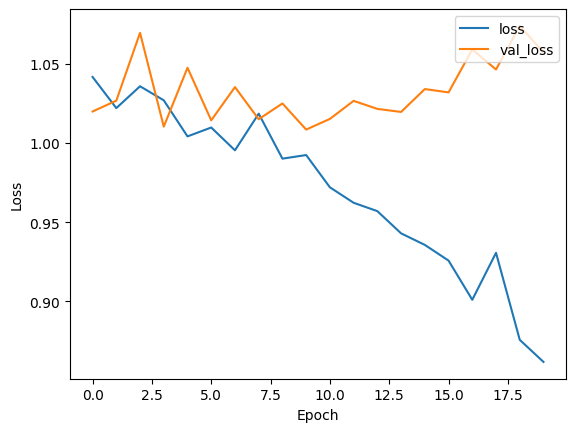}
         \caption{Loss for SciBERT model}
         \label{fig:three sin x}
     \end{subfigure}
        \caption{Accuracy and loss for the SciBERT model for decision tree, SVC, random forest, and logistic regression classifiers}
        \label{fig:three graphs}
\end{figure}

\begin{figure}[ht!]
     \centering
     \begin{subfigure}[b]{0.40\textwidth}
         \centering
         \includegraphics[width=\textwidth]{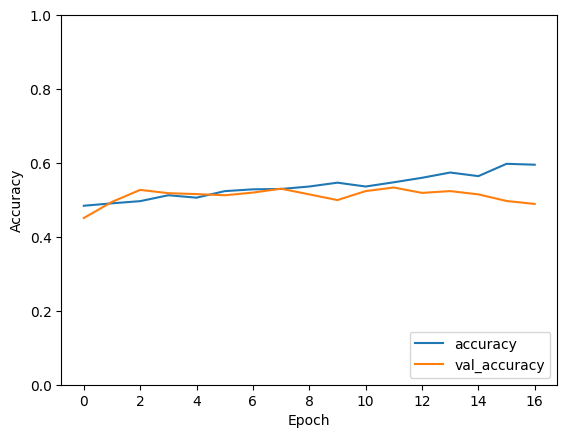}
         \caption{Accuracy for BioBERT model}
         \label{fig:y equals x}
     \end{subfigure}
     \hfill
     \begin{subfigure}[b]{0.40\textwidth}
         \centering
         \includegraphics[width=\textwidth]{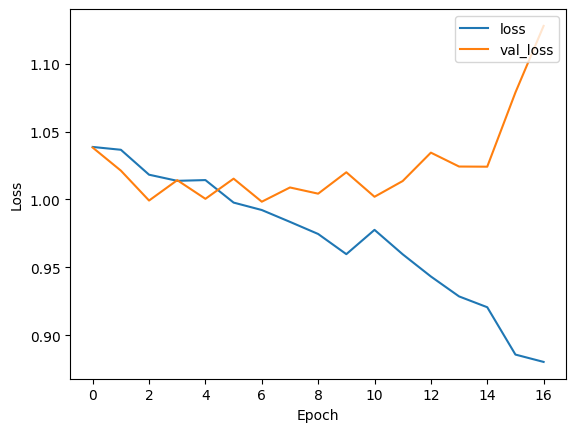}
         \caption{Loss for BioBERT model}
         \label{fig:three sin x}
     \end{subfigure}
        \caption{Accuracy and loss for the BioBERT model for decision tree, SVC, random forest, and logistic regression classifiers}
        \label{fig:three graphs}
\end{figure}
\clearpage
\section{Conclusion and Future Work}
This work proposed a machine learning and natural language processing approach for sentiment classification of drug reviews created and posted by users. Analyzing such publicly available information is crucial to finding out the hidden patterns in drug usage and any adverse drug reactions that will be of interest to different stakeholders for an informed decision-making process. This work uses advanced pre-trained models for feature extraction from publicly available user reviews on drugs, and different machine learning classifiers are trained for the classification task. A comparative analysis of the precision, recall, and f1-score for BERT, SBERT, BioBERT, and SciBERT is performed and the results are reported. As the results are promising, one future research dimension is to train advanced deep learning models on a large number of data points to improve classification accuracy.

\bibliographystyle{unsrtnat}
\bibliography{references} 
\end{document}